\documentclass[10pt,twocolumn,letterpaper]{article}

\usepackage{iccv}
\usepackage{times}
\usepackage{epsfig}
\usepackage{graphicx}
\usepackage{amsmath}
\usepackage{amssymb}
 
\usepackage{algorithm}
\usepackage[noend]{algorithmic}
\usepackage{bm}
\usepackage{float}

\usepackage[pagebackref=true,breaklinks=true,letterpaper=true,colorlinks,bookmarks=false]{hyperref}

\usepackage[breaklinks=true,bookmarks=false]{hyperref}

\iccvfinalcopy 

\def\iccvPaperID{****} 

\ificcvfinal\fi

\begin{document}

\title{Task-Attentive Transformer Architecture for Continual Learning of Vision-and-Language Tasks Using Knowledge Distillation}

\author{Yuliang Cai\\
University of Southern California\\
{\tt\small caiyulia@usc.edu}
\and
Jesse Thomason\\
University of Southern California\\
{\tt\small jessetho@usc.edu}
\and
Mohammad Rostami\\
University of Southern California\\
{\tt\small rostamim@usc.edu}
}

\maketitle
\ificcvfinal\thispagestyle{empty}\fi

\begin{abstract}

The size and the computational load of  fine-tuning  large-scale pre-trained neural network are becoming two major obstacles in adopting machine learning in many applications. Continual learning (CL) can serve as a remedy through enabling knowledge-transfer across sequentially arriving tasks which relaxes the need to fine-tune all network weights from scratch. However, existing CL algorithms primarily consider learning unimodal vision-only or language-only tasks. We develop a transformer-based CL architecture for learning bimodal vision-and-language tasks based on increasing the number of the learnable parameters  dynamically and using knowledge distillation. The new additional parameters are used to specialize the network for each task. Our approach enables sharing information between the tasks while addressing the challenge of catastrophic forgetting. Our approach is scalable learning to a large number of tasks because it requires little memory and time overhead. Our model reaches state-of-the-art performance on  challenging vision-and-language tasks.

\end{abstract}

\section{Introduction}

Large-scale pre-trained transformer models are increasingly replacing prior deep learning architectures in a wide range of applications and modalities,  including vision-and-language task ~\cite{kenton2019bert,dosovitskiyimage,kim2021vilt}.
These models are usually pretrained on a   large dataset and then are fine-tuned to generalize on a downstream task. Not only, fine-tuning compromises the model generalizability, it necessitates storing a copy of the base model for each task.
Continual learning (CL) algorithms~\cite{jin2021learn,yang2022continual,wang2022continual,pelosin2022towards,ermis2022continual} have explored mitigating these challenges for transformers through using a shared model that benefits from cross-task knowledge transfer while     overcoming  catastrophic forgetting effects~\cite{kirkpatrick2017overcoming}. However, current CL methods   consider unimodal tasks, e.g., vision-only~\cite{lin2021clear,https://doi.org/10.48550/arxiv.2111.11326} or language-only tasks~\cite{jin2021learn,yang2022continual}, and learning sequentially arriving multimodal tasks using a shared architecture is an extremely under-explored area.

CL algorithms use three primary strategies to alleviate catastrophic forgetting. A group of CL algorithms regularize a fixed shared model to learn each task through different information pathways, e.g., weights~\cite{kirkpatrick2017overcoming,aljundi2018memory,rostami2018multi}. The core idea is to identify a subset of model parameters that are important to encode the learned knowledge about each task and then consolidate these parameters when updating the model to learn new tasks. Another group of algorithms  are based on   model expansion~\cite{rusu2016progressive,yoonlifelong}. The idea is to expand a base model via a small number of additional weights and specialize the network to learn new tasks through these   weights. In these approaches, the shared weights enable   knowledge transfer across similar tasks. Finally, a group of algorithms use   pseudo-rehearsal through experience replay ~\cite{rolnick2019experience,rostami2021lifelong}. The idea is to store a representative subset of training data for each task and store then in a memory buffer, and replay back those samples along with the current task's data to maintain the encoded knowledge about the past   task. Some methods relax the need for a memory buffer by enabling the model to generate pseudo-samples for the past learned tasks that are used to implement experience replay~\cite{shin2017continual,rostami2020generative}.  

In this paper, we develop a new algorithm for learning vision-and-language tasks in a CL setting based on dynamic model expansion. To this end, we leverage the bimodal ViLT transformer~\cite{kim2021vilt}. The self-attention layers of ViLT are used as a shared encoder 
across all tasks. We equip the base ViLT model with task-attention layers \cite{https://doi.org/10.48550/arxiv.2111.11326} that help to specialize the model for each task using a task-specific token.
To the best of our knowledge, we are the first to develop a vision-and-language transformer architecture for multimodal CL. 
Our architecture leads to a small memory overhead and a limited inference time overhead during testing. It also does not need extensive hyper-parameters and remains robust when facing  an unknown number of tasks.

 Our specific contributions include:
\begin{itemize}
    \item  A dynamically expanding transformer architecture for multimodal CL, the first of its own kind, which requires only limited extra storage for each new task. 
    \item  The training algorithm that can handle diverse tasks, in a sequence, with different vision-and-language tasks.
    \item  Extensive experiments to demonstrate that our architecture achieves state-of-the-art performance  on sequential vision-and-language (VaL) tasks.
\end{itemize}

\section{Background and Related Work}

Our work is the first architecture of its own kind to enable multimodal continual learning using transformers.
 
\paragraph{Transformers:}
The first transformer architecture was developed  for machine translation. It is based on encoder and decoder layers that implement the   self-attention mechanism~\cite{vaswani2017attention}. 
The self-attention layers  implement an attention mechanism on an input sequence by relating the positions of tokens in the sequence to the way that a global feature vector is extracted from the sequence. 
As a result, the effect of some tokens  is enhanced while other tokens are overlooked in the extracted representations, leading to attending to a small but informative number of tokens of the input sequence.
The idea of self-attention then was adopted for other modalities such as computer vision tasks~\cite{dosovitskiy2020image} based on modeling an image using a sequence of  patches as tokens. These patches result from the decomposition of the input image into blocks. Various improved transformer architectures are developed by modifying the base idea and training procedures for each language and image modalities.

More recently, bimodal transformers are developed for processing vision-and-language tasks~\cite{su2019vl,tan2019lxmert,kim2021vilt,viltbert,chen2020uniter}. The core idea is to use modality-specific self-attention layers to extract suitable features from each of the vision and language inputs.  These features then are integrated at higher layers to extract cross-modal contextualised representations of the
multimodal inputs. The idea is that these global vectors can  model the interaction between the vision and the language inputs which is necessary to perfrom VaL tasks well.
The idea has also been adopted in other modalities, including in the context of video processing to relate vision and speech inputs~\cite{arnab2021vivit,sun2019videobert}.
These transformers are usually trained on a suitable large-scale dataset and 
  have been found to be highly effective when
fine-tuned on downstream tasks. Due to a significant performance improvement, transformers are   replacing older architectures in most modalities.

\paragraph{Continual learning for transformers:}

Despite the successful adoption of transformers   in various benchmarks, fine-tuning naturally compromises their generalizability. Using an independent transformer per task would lead to a significant memory load as the size of transformers are increasing dramatically to account for solving more abstract tasks. CL seems to be a natural solution for these challenges but
works on CL using transformers are   limited. 
Xin et al.~\cite{jin2021learn} use adapters in combination with a hypernetwork to enable CL for language tasks. Adapters weights are generated by the hypernetwork which are placed between frozen layers of a language transformer to make it adaptive. Alternatively, Yang et al.~\cite{yang2022continual} propose a transformer calibration module to make a transformer   adaptive. The calibration module is considered to be independent from the base pre-trained transformer and helps to specialize it on a downstream task.
The Lifelong Vision Transformer \cite{wang2022continual} utilizes an inter-task attention mechanism to integrate information from previous tasks and reduces the rate at which important attention weights shift away from old tasks towards the current task. 
 Douillard et al.~\cite{https://doi.org/10.48550/arxiv.2111.11326} propose a CL architecture for vision tasks using the ViLT model~\cite{kim2021vilt}. Pelosin et al.~\cite{pelosin2022towards} extend this work to   an exemplar-free setting via distilling the attention-level matrices of transformers to enable model plasticity and to mitigate forgetting effects. Ermis et al.~\cite{ermis2022continual} use the idea of adapters in a vision context.
To the best of our knowledge, no prior work has explored CL for multimodal tasks using transformer architectures.

\begin{figure}
\includegraphics[scale=0.45]{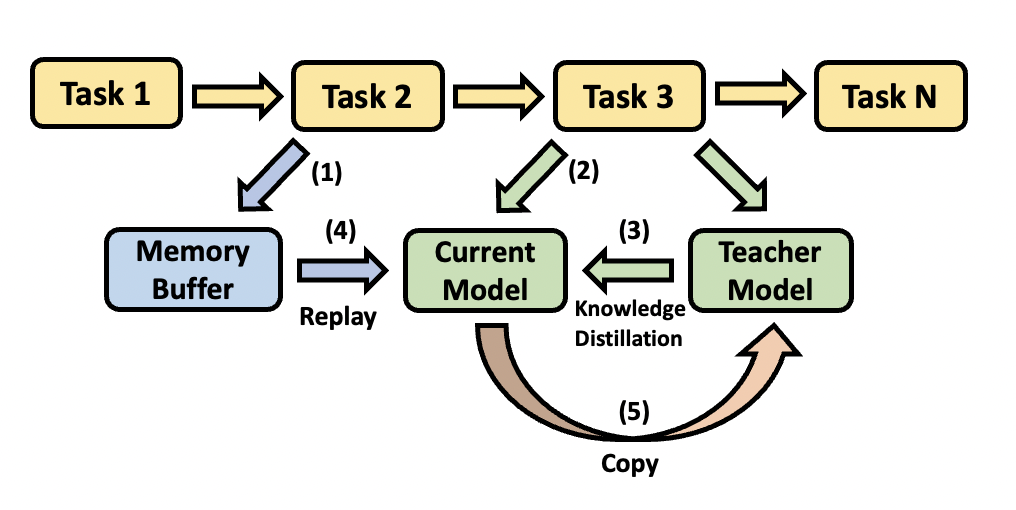}
\caption{\small The proposed  CL training procedure for each task: (1) A small portion of the data for previous tasks are randomly selected and stored in a memory buffer. (2) The current task  arrives with $\mathcal{D}^i$. (3) The training data $\mathcal{D}^i$ is used  as input to the teacher model to compute the knowledge distillation loss. (4) The memory buffer samples are replayed along with the current task data to train the main model. (5) After learning the current task,   the teacher model of the next task would be a copy of the current main model.}
\label{fig:Diagram}
\end{figure}

\begin{figure*}
\includegraphics[scale=0.6]{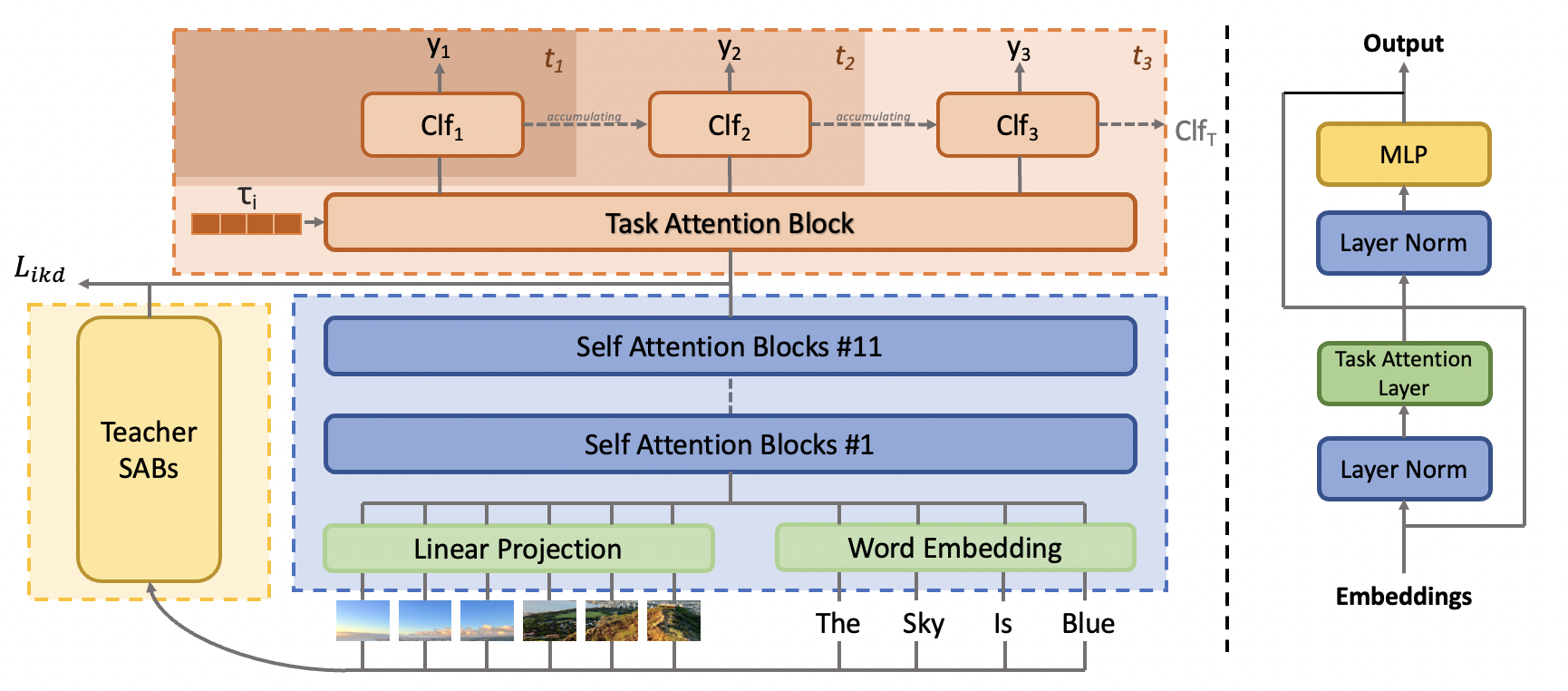}
\label{fig:Big_Diagram}
\caption{The proposed transformer-based architecture: (left) The VaL inputs are converted into two sequences and then fed into the self-attention layers to generate a fused global feature vector. The data feature vector is then concatenated with the learnable task-specific tokens and then fed into the task attention layer to generate the input for the task-specific classifier heads. The same VaL inputs are also fed into the teacher model's transformer architecture to compute Knowledge Distillation. (right) The task-attention block architecture.}
\end{figure*}
\section{Problem Description}
Consider a set of sequentially arriving VaL tasks $\{\mathcal{T}_i\}_{i=1}^T$, each with the annotated training dataset $\mathcal{D}^i$ = $\{\langle(\bm{I}_i^j,\bm{L}_i^j)^i, y^j_i\rangle_{j=1}^{N_i}\}$, where $\bm{I}_i^j \in \mathbb{R}^{H\times W\times C}$ represents the image input,  $\bm{L}_i^j \in \mathbb{R}^{L\times |V|}$ represents the language input, while $y^j_i$ is the text-typed discrete label. The order of these tasks and $T$ are not known a priori. The training data points for $\mathcal{T}_i$ are assumed to be drawn iid from a task-specific joint distribution $p_i^t(\cdot,\cdot,\cdot)$. our goal in multimodal CL is to learn each task at time-step $i$ and then move forward to learn the next tasks. The learned tasks can be encountered at any time during testing in the future and hence, we would like to maintain the ability to perform the learned tasks well.

When learned in isolation, each of these bimodal tasks $\mathcal{T}_i$ can be learned using supervised learning conditioned on selecting the suitable predictive model $f^i_{\theta_M}(\cdot,\cdot)$, e.g., a transformer  with trainable parameters $\theta_M$, and the discrimination loss   $\mathcal{L}(\cdot)$, e.g., cross-entropy. 
However, Due to the storage limit, we assume  a shared model should be used which makes single task learning an impractical solution. Additionally,   the training data for each task $\mathcal{T}_i$ is considered to be inaccessible after learning that task. As a result,   multitask learning~\cite{caruana1998multitask} would be an impractical solution as well.
The single task learning strategy using a shared model is not applicable in CL either.  Because when the model is updated to learn the current tasks, its performance on the past learned tasks will degrade due to catastrophic forgetting~\cite{kirkpatrick2017overcoming}. 


Figure~\ref{fig:Diagram} visualizes the high-level presentation of the solution that we propose to address multimodal CL.
To use a shared model across all tasks and benefit from knowledge transfer across all tasks, we consider a base transformer model as  $f_\theta(\cdot,\cdot)$ and make it adaptive by adding a unique task attention layer after its final layer. Additionally, we modify the loss supervised learning loss by adding a knowledge distillation loss~\cite{hinton2015distilling} on the intermediate model layers. We consider the teacher model in the knowledge distillation formulation to be a copy of $f^{i-1}_{\theta}(\cdot,\cdot)$ when training the main student model on $i^{th}$ task $\mathcal{T}_i$. Additionally, we rely on pseudo-rehearsal through experience replay \cite{rolnick2019experience} using a small memory buffer to tackle catastrophic forgetting. 

\section{The Proposed Architecture}
Figure~\ref{fig:Big_Diagram} visualizes our transformer-based architecture for multimodal CL.  The architecture consists of a shared pre-trained replaceable bimodal transformer, an independent task attention block, and MLP classification headers. The task attention block receives task-specific tokens that make the model adaptive. We provide details about these modules and the strategies that we use to train it.

\subsection{Sequential Feature Extraction Block}
We follow the ViLT feature generation procedure~\cite{kim2021vilt}. To build a sequence from the input images, we decompose a given image  $\bm{I} \in \mathbb{R}^{H\times W \times C}$ into patches and then flatten these patches to generate 2D vectors $\bm{v} \in \mathbb{R}^{N\times (P^2 \cdot C)}$. Here, $C$ is the number of channels, $P \times P$ is the size of each image patch, and $N = HW/P^2$ is the number of patches with size $P \times P$. After generating the set of vectors, we apply a trainable linear projection, $\bm{V} \in \mathbb{R}^{(P^2 \cdot C) \times H}$, and a position embedding, $\bm{V}^{pos} \in \mathbb{R}^{(N+1)\times H}$, to transform $\bm{v}$ into the sequential representation $\bm{\overline{v} }\in \mathbb{R}^{N\times H}$\:
\begin{equation}
    \bm{\overline{v}} = [\bm{v}^{class};\bm{v}_1\bm{V};\bm{v}_2\bm{V};...;\bm{v}_N\bm{V}] + \bm{V}^{pos}
\end{equation}

We generate word vectors for the language input $\bm{l} \in \mathbb{R}^{L\times |V| }$ after applying a word embedding matrix $\bm{T} \in \mathbb{R}^{|V|\times H}$ and using a position embedding matrix $\bm{T}^{pos} \in \mathbb{R}^{(L+1)\times H}$, we embed data into $\bm{\overline{t}} \in \mathbb{R}^{L\times H}$.
\begin{equation}
    \bm{\overline{t}} = [\bm{t}^{class};\bm{l}_1\bm{T};\bm{l}_2\bm{T};...;\bm{l}_L\bm{T}] + \bm{T}^{pos}
\end{equation}
We then sum the image and text embeddings with their independent model-type embedding vector $\bm{t}^{type}$ and $\bm{v}^{type} \in \mathbb{R}^H$, and   concatenate them to build a single sequence $\bm{s}$. 
\begin{equation}
    \bm{s^0} = [\bm{v^{type}}+\bm{\overline{v}};\bm{t^{type}}+\bm{\overline{t}}]
\end{equation}
The combined vector $\bm{s}$ is passed through   $D$ classic self-attention layers of the base bimodal transformer, i.e., ViLT, and then are fed into the task attention layer, with the output embedding from the transformer attention layers denoted as $\bm{s}^D$ and output from the task attention layer   as $\bm{s}^{D+1}$.
\begin{equation}
\begin{aligned}
    &\bm{\hat{s}}^{d} = MSA(LN(\bm{s}^{d-1}))+\bm{s}^{d-1},    d = 1,...,D \\
    &\bm{s}^{d} = MLP(LN(\bm{\hat{s}}^{d}))+\bm{\hat{s}}^{d},   d = 1,...,D
\end{aligned}
\end{equation}

\subsection{Task Attention Block}
\label{4.2}
The core idea of our work lies in adopting the idea of self-attention on task-level, each time a new task is learned. Different from the vanilla input data level self-attention layer, the task-attention layer is a task-focused attention layer that generates a task token for each new task, denoted as $\tau_i \in \mathbb{R}^{G\times 1}$ for task $i \subseteq [1,2,..,T]$, where $G$ is the size of latent space of each self-attention layer. Similar to the self-attention block (SAB), the task attention block (TAB) is a module which consists of an attention layer, layer normalization, and MLP, while the self-attention layer is replaced by the task-attention layer. Task attention block takes two inputs, the task token $\tau$ and the output of the self-attention blocks $\bm{s}^D$. The two vectors are concatenated to generate a single vector as the task attention input:
\begin{equation}
    \begin{aligned}
    &\bm{s}{'}_i^{D+1} = [\tau_i, \bm{s}^{D}] \in \mathbb{R}^{(N+1)\times G}, i = 1,...,T\\
        &\bm{\hat{s}}_i^{D+1} = TA(LN(\bm{s}{'}_i^{D+1})), i = 1,...,T \\
&\bm{s}_i^{D+1} = MLP(LN(\bm{\hat{s}}_i^{D+1})) + \bm{\hat{s}}_i^{D+1}, i = 1,...,T\\
    \end{aligned}
\end{equation}
The task attention block is placed after the last self-attention block of the transformer. While we can have more than one task attention  block, our architecture uses a single TAB.

The operation of the task attention block is given below:
\begin{equation}
\begin{aligned}
&\bm{Q}_i = \bm{W}_q \times \tau_i, \\
&\bm{K}_i = \bm{W}_k \times \bm{s}_i^{D+1}, \\
&\bm{V}_i = \bm{W}_v \times \bm{s}_i^{D+1}, \\
&\bm{A}_i = Softmax(\bm{Q}_i \cdot \bm{K}_i^T/\sqrt{G/h}), \\
&\bm{O}_i = \bm{W}_o\bm{A}_i\bm{V}_i + \bm{b}_o \in \mathbb{R}^{1 \times G}
\end{aligned}
\end{equation}
where $h$ is the number of attention heads \cite{https://doi.org/10.48550/arxiv.1706.03762}. 

Finally, the output of the task attention block, $\bm{s}_i^{D+1}$ is then fed into task-specific classifier layers:
\begin{equation}
    \bm{y}_i = Clf_i(\bm{s}_i^{D+1}), i = 1,...,T
\end{equation}

\section{Training Algorithm}
The architecture in Figure~\ref{fig:Big_Diagram} visualizes a snapshot of our model at a timestep but this architecture is dynamically being expanded as more tasks are learned. We also need a suitable loss function to train in a CL setting.

\subsection{Token Expansion}
During the training stage, the transformer's self attention blocks and the task attention block are shared among all the tasks. However, for each of the new tasks, we define a new task token with the same dimension, $\tau \in G\times 1$, and initialize a new task-specific classifier, $Clf_i(\cdot)$ for task $i$. With more tasks added on, the output dimension of task-specific classifier $i$ would be accumulating:
\begin{equation}
    \bm{E}_i = \bm{E}^{orig}_i + \bm{E}_{i-1}, i = 1,\ldots,T,
\end{equation}
Where $\bm{E}_i$ denotes the output dimension for $i^{th}$ classifier, $\bm{E}^{orig}_i$ denotes the output dimension for $i^{th}$ task in its original design.
For task $i$, we combine the $i^{th}$ task token with the updated path token, $\bm{s}^D$ from the last self-attention block of the transformer, and send it into the task attention block, as described in Sec \ref{4.2}. At this stage, only the $i^{th}$ task token and task-specific classifier would be trainable, while all the other task tokens and classifiers remain frozen. 

During the test stage, the task number, $i$, of test data is explicitly given, and $\bm{s}^D$ is combined with the $i^{th}$ learned task token to feed into task attention block along with using its corresponding task-specific classifier, while all other the task tokens and classifiers remain unused. 

\subsection{Loss and Knowledge Distillation}
Our objective function consists of three loss terms: (i) cross-entropy Loss, $\mathcal{L}_c$, which is the original objective function for each task in single task learning setting. Note it can vary from task to task, (ii) the knowledge distillation (KD) loss, which is computed from $\bm{s}^D$ of the main student model and the teacher model, $\mathcal{L}_{ikd}$, (iii) the diverse loss $\textit{L}_{div}$ which compares the data distribution of task tokens and makes them more diverse. The final loss is as follows:
\begin{equation}
    \mathcal{L} = (1-\lambda)\mathcal{L}_c + \lambda\alpha\mathcal{L}_{ikd} + \beta\mathcal{L}_{div},
\end{equation}
where $\lambda$ is set to $\frac{T_n-1}{T_n}$, $T_n$ is the total number of tasks we have seen so far, $\alpha$ is a constant which varies for different tasks, and 
$\beta = min(\mathcal{L}_{div}, 0.1\times ((1-\lambda)\mathcal{L}_c + \lambda\alpha\mathcal{L}_{ikd}))$.
 
The application of intermediate knowledge distillation is the core of our design, which aims to distill the knowledge from the teacher model into the main student model in order to constrain the distribution shift and prevent catastrophic forgetting.
In our architecture, the teacher model is a copy of model $f^{i-1}_{\theta}(\cdot,\cdot)$ when training on $i^{th}$ task. Different from most other methods which apply knowledge distillation by computing the loss from the last layer output, $y$ and $y_{teacher}$, we introduce an intermediate knowledge distillation loss, i.e., the loss term $\mathcal{L}_{ikd}$ is computed between the last self-attention block of the transformer and the task attention block, $\bm{s}^{D}$. Through experiments, we find that compared with knowledge distillation computed from the last layer output, $y$ and $y_{teacher}$, such an intermediate knowledge distillation is helpful in the architecture of a pre-trained model is followed by a non-pre-trained block, which could constrain the probability shift of pre-trained parameters and leave the rest of the layers flexible enough to learn new tasks. To our best knowledge, we are the first method that introduces intermediate knowledge distillation objective function in the multimodal CL field.

\subsection{Experience Replay}

The above training procedure allows training a shared model across the tasks but still, we need datasets for all tasks which breaks a primary assumption in CL. To address this issue, we use a memory buffer
during the training stage at each time-step   which stores a tiny percentage, e.g., $\approx 1\%$, of the training dataset for all the previous tasks~\cite{rostami2021detection}. When learning the current task with a specific batch number, the next batch will be randomly selected from the memory buffer to consolidate the parameter distribution on previous tasks.
As a result, forgetting effects will be mitigated.

The training procedure for our algorithm, named Task Attentive Multimodal Continual Learning (TAM-CL), is presented in Algorithms 1 and 2.

\begin{algorithm}
\caption{TAM-CL TrainStep}\label{alg:step}
\begin{algorithmic}
\STATE\textbf{INPUT:} Model \textbf{M}, Teacher Model \textbf{T}, Batch, Target \textbf{t}, Token \textbf{k}
\STATE p $\leftarrow$ Model(Batch)
\STATE $loss$ $\leftarrow$ CrossEntropyLoss(p, \textbf{t})
\STATE ModelSabPre $\leftarrow$ \textbf{M}.SAB(Batch)
\STATE TeacherSabPre $\leftarrow$ \textbf{T}.SABs(Batch)
\STATE $loss_{ikd} \leftarrow$ KL-Div(ModelSabPre, TeacherSabPre)
\STATE $loss_{div} \leftarrow$ CrossEntropyLoss($\textbf{k}_i,\textbf{k}_j)\quad j=1,..,i-1$\\
$loss$ $\leftarrow$ $(1-\lambda)loss$ + $\lambda\alpha loss_{ikd}$ + $\beta loss_{div}$\\
 $loss$.backward()\\
\textbf{Return} loss
\end{algorithmic}
\end{algorithm}

\begin{algorithm}
\caption{TAM-CL Train}
\begin{algorithmic}
\STATE \textbf{INPUT:} Model \textbf{M}, MemBuffer \textbf{B}, ReplayFreq \textbf{f}\\
\FOR{epoch in num\_epoch}
\FOR{step, batch in dataloader}
\STATE Loss $\leftarrow$ TrainStep(\textbf{M}, batch)
\IF{step \% \textbf{f} == 0}
\STATE $\text{batch}_{replay}$ $\leftarrow$ \textbf{B}.getBatch()
\STATE $\text{Loss}_{replay}$ $\leftarrow$
TrainStep(\textbf{M}, $\text{batch}_{replay}$)
\ENDIF
\ENDFOR
\ENDFOR
\end{algorithmic}
\end{algorithm}
\section{Experimental Results}
Inspired by the   CLIMB benchmark for multimodal CL~\cite{srinivasanclimb}, we evaluate  TAM-CL using four VaL benchmarks.

\begin{table}
\small
\centering
\begin{tabular}{||c|c|c|c||}
    \hline
    Name & \# Images & \# Txt-Img pairs & \# Labels\\ \hline
    SNLI-VE & 29783 & 529527 & 3\\ 
    COCOQA & 123287 & 78736 & 430\\ 
    PathVQA & 2499 & 17325 & 4092\\ 
    NLVR2 & 107292 & 86373 & 2\\
    \hline
\end{tabular}
\caption{Statistics of the VaL dataset.}
\label{tab:dataset}
\end{table}

\subsection{Experiment Setup}
\label{sec:6.1}
We use four diverse VaL datasets to generate sequential tasks. We use \textbf{SNLI-VE}\cite{https://doi.org/10.48550/arxiv.1901.06706}, an image-sentence pairs dataset whereby a premise is defined by an image, rather than a natural language sentence as in traditional Textual Entailment tasks, \textbf{COCOQA}\cite{https://doi.org/10.48550/arxiv.1505.02074}, a visual question answering dataset based on Microsoft COCO image dataset, \textbf{PathVQA}\cite{https://doi.org/10.48550/arxiv.2003.10286}, a pathology VQA dataset of which the images are all chosen from the medical field, \textbf{NLVR2}\cite{https://doi.org/10.48550/arxiv.1811.00491}, a visual reasoning dataset which takes two images as input and determine the correctness of the given sentence. Table~\ref{tab:dataset} provides statistical details of these dataset.

Although our approach is applicable to common VaL transformer architectures, we use ViLT model with pre-trained weights, ``BERT-base-uncased'' in our experiments. 
 To maintain consistency in our comparison, we use ViLT   with the same pre-trained parameters for all the experiments. Hence, we have 11 self-attention blocks (SAB), with dimension of 768 and attention heads of 12. We then attach one task attention block (TAB) after the transformer encoder which also has 768 hidden dimensions and 12 attention heads. For all four tasks, we apply AdamW optimizer with $l$ = 1e-2, $\epsilon$ = 1e-8, $\beta_1$ = 0.9, and $\beta_2$ = 0.98.

Since there is no prior method for multimodal CL, we use EWC, Experiment Replay, and direct fine-tuning, as three primary alternatives for comparison. Direct fine-tuning serves as a lowerbound to measure the effect of catastrophic forgetting. Improvements over EWC and Experiment Replay
which are two primary CL approaches demonstrate that our algorithm is effective.

After learning each task $\mathcal{T}_i$, we evaluate the forgetting rate on previous tasks $T_k$, $k \in 1,..,i-1 $. 
 To study the effect of task order, we  perform   experiments with different task orders that reflect the difficulty of tasks. For each of the learned tasks, the performance accuracy and the forgetting rates on all previous tasks are reported.  
We rank the difficulty level of the four tasks by an intuitive metric: $\frac{\#\ Txt-Img\ Pairs}{\#\ Labels}$. The bigger this value is, the easier the tasks is. In this case, we rank our four tasks from difficult to easy as: \textbf{PathVQA}, \textbf{COCOQA}, \textbf{NLVR2} and \textbf{SNLI-VE}, of which corresponding scores are 4.23, 183.11, 43186.50 and 176509.00. To eliminate the bias brought by a specific task order, we perform   experiments om several different task orders, including from the most difficult to the easiest, from the easiest to the most difficult, and two   random task orders.

Most CL algorithms consider relatively homogeneous tasks.
To evaluate the algorithm capacity for preventing catastrophic forgetting on heterogeneous tasks that we have, we use the following normalization-based metric~\cite{srinivasanclimb}:
\begin{equation}
    \mathbb{T}_\textbf{F}(j \leftarrow i) = \frac{S^j_A-S^{j\leftarrow i}_A}{S^j_A-S^j_R} 
    \label{eqnor}
\end{equation}
where $\mathbb{T}_\textbf{F}(j \leftarrow i)$ stands for the forgetting rate of task j after learning task i, $S^j_A$ denotes the accuracy of task $j$ before learning new tasks, $S^{j\leftarrow i}_A$ denotes the accuracy of task $j$ after learning task $i$, $i > j$, and $S^j_R$ means the accuracy of task $j$ by randomly choosing the output label, which is calculated by $\frac{1}{\#\ labels}$. In other words, Eq.~\eqref{eqnor} enables forgetting rates across tasks that are considerably different because we are measuring how well the model is preforming compared to a baseline of total forgetting for that task.  More experimental details are included in the Appendix.


\begin{table*}[h]
\begin{tabular}{||c|c|c|c|c|c|c||}
    \hline
    \multicolumn{7}{||c||}{COCOQA $\rightarrow$ NLVR2 $\rightarrow$ PathVqa $\rightarrow$ SNLI-VE} \\ \hline
    & NLVR2 & \multicolumn{2}{|c|}{PathVqa} & \multicolumn{3}{|c||}{SNLI-VE} \\ \hline
    & COCOQA & COCOQA & NLVR2 & COCOQA & NLVR2 & PathVqa \\ \hline
    TAM-CL & \textbf{11.68\%(66.62)} & \textbf{91.28\%(6.58)} & \textbf{95.51\%(50.92)} & \textbf{85.97\%(10.58)} & \textbf{93.92\%(51.36)} &  \textbf{10.19\%(49.69)}\\ 
    Finetune & 97.06\%(1.60) & 99.43\%(0.42) & 104.51\%(49.18) & 99.11\%(0.66) & 102.30\%(49.58) & 71.34\%(15.24)\\ 
    EWC & 97.71\%(0.94) & 99.40\%(1.21) & 100.56\%(50.79) & 99.25\%(0.57) & 95.26\%(50.87) & 32.99\%(36.47)\\ 
    ER & 31.00\%(51.87) & 97.64\%(3.94) & 105.10\%(48.95) & 86.92\%(9.83) & 98.18\%(50.37) & 17.84\%(43.80)\\
    \hline
    \multicolumn{7}{||c||}{SNLI-VE $\rightarrow$ NLVR2 $\rightarrow$ COCOQA $\rightarrow$ PathVqa} \\ \hline
    & NLVR2 & \multicolumn{2}{|c|}{COCOQA} & \multicolumn{3}{|c||}{PathVqa} \\ \hline 
    & SNLI-VE & SNLI-VE & NLVR2 & SNLI-VE & NLVR2 &  COCOQA\\ \hline
    TAM-CL & \textbf{17.57\%(68.38)} &  \textbf{35.74\%(60.66)} & \textbf{41.06\%(61.04)} & \textbf{94.79\%(35.55)} &  \textbf{95.49\%(50.85)} &  \textbf{92.43\%(4.88)}\\ 
    Finetune & 75.84\%(43.52) & 98.75\%(33.86) & 94.22\%(50.99) & 96.02\%(35.01) & 95.87\%(50.80) & 97.98\%(1.26)\\ 
    EWC & 72.00\%(45.20) & 97.15\%(34.54) & 92.73\%(51.26) & 99.09\%(33.72) & 96.45\%(50.66) & 97.55\%(1.41)\\ 
    ER & 43.73\%(57.22) & 61.38\%(49.73) & 63.69\%(56.62) & 101.72\%(32.60) & 97.61\%(50.43) & 93.15\%(4.32)\\
    \hline
\end{tabular}
\caption{The forgetting rate for each task in two different task sequences. For each task sequence, the task names on the first row represent the current tasks, and the task names on the second row represent the previous tasks. Each value means the \textbf{forgetting rate\% (current accuracy)} of the previous task after learning the current task.}
\label{tab:comparative}
\end{table*}

\subsection{Comparative Results}

We report  the performance results of the algorithms in Table \ref{tab:dataset}. We have reported  the performance results for two task sequences. As expected,   we   observe that fine-tuning is the worst algorithm for most of the accuracy and forgetting rates at different time-steps. This empirical observation demonstrates  the significance of adopting CL algorithm for learning multimodal tasks sequentially. We also observe that EWC has  similar performances to those of direct fine-tuning.
EWC often is effective when used on smaller models but in line with prior observations in the case of using transformers for unimodal tasks~\cite{srinivasanclimb,jin2021learn,https://doi.org/10.48550/arxiv.2111.11326}, we
conclude that  regularization-based CL methods such as EWC are not suitable methods for CL with large transformer-based models. This result suggests that transformers are sensitive with respect to weight consolidation because their learning capacities are compromised significantly.
In contrast, the accuracy and forgetting rate results for experience replay method is decent, especially when learning early tasks. For example, in the first task sequence in Table \ref{tab:comparative}, the forgetting rate of COCOQA after learning NLVR2 is 31.00\%, compared with 97.71\% by EWC and 97.06\% by fine-tuning. In the second task sequence, the forgetting rate of SNLI-VE after learning NLVR2 is 43.73\% by ER, while for fine-tuning and EWC, the forgetting rates are both above 70\%.

Finally, we observe that for TAM-CL and in the task sequence COCOQA $\rightarrow$ NLVR2 $\rightarrow$ PathVqa $\rightarrow$ SNLI-VE, the forgetting rate of COCOQA after learning NLVR2 is 11.68\%, which is significantly  better than the other three methods. After learning the PathVqa dataset, the forgetting rate of COCOQA is 91.28\%, better than  experience replay  which achieves 97.64\%. The forgetting rate on NLVR2 is 95.51\% which is better than EWC by a margin of 5.05\%. When SNLI-VE learned, TAM-CL not only performs well in remembering the early tasks, with forgetting rates of 85.97\% and 93.92\%, respectively, but gets 10.19\% forgetting rate on PathVqa which is the most difficult task among the four tasks. In task sequence SNLI-VE $\rightarrow$ NLVR2 $\rightarrow$ COCOQA $\rightarrow$ PathVqa, our forgetting rate of NLVR2 on SNLI-VE is also leading, which is 26.16\% higher than the $2^{nd}$ top method. After learning the third task, TAM-CL's forgetting rate is lower than the $2^nd$ top method by 25.64\% and 22.63\% for task 1 and task 2 ,respectively. After learning the $4^{th}$ task, the overall forgetting rates of the four methods increase to a relatively high level for all methods, but TAM-CL still outperforms the other methods. 

\begin{figure}
\includegraphics[scale=0.25]{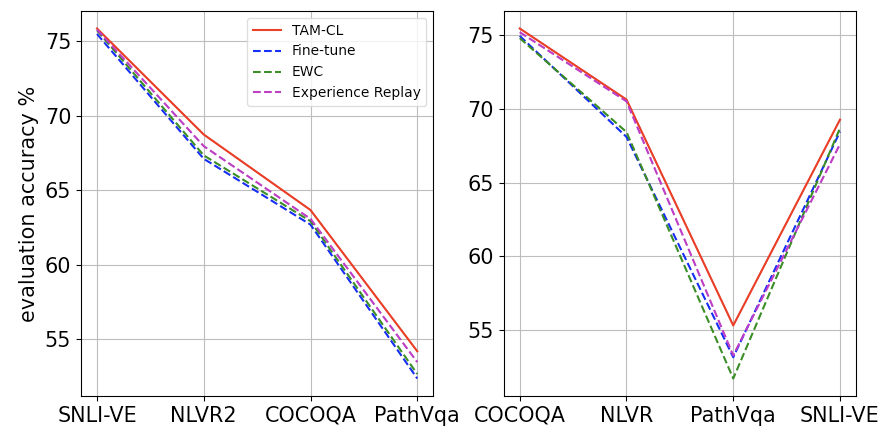}
\caption{\small Accuracy learning curves for two different   sequences.}
\label{fig:comparison}
\end{figure}

Meanwhile, although the primary focus of CL is on preventing catastrophic forgetting, the accuracy learning curves presented in Figure \ref{fig:comparison} demonstrate that our method also enables forward transfer, i.e., past experiences help the model to learn the current task more efficiently compared to learning it directly using the pre-trained model, because similarities between the tasks can help to share knowledge. We conclude that distillation using the teacher model is enables transferring the learned knowledge fro past tasks well, particularly, using the intermediate layers that encode high-level sharable knowledge. We observe that for the two task sequences, TAM-CL outperforms the alternatives both in terms of backward transfer, i.e., catastrophic forgetting, and forward transfer, i.e., cross-task knowledge transfer. Additional experiments are included the Appendix.

\subsection{Ablative Results}

\begin{table*}[h]
\begin{tabular}{||c|c|c|c|c|c|c||}
    \hline
    \multicolumn{7}{||c||}{COCOQA $\rightarrow$ NLVR2 $\rightarrow$ PathVqa $\rightarrow$ SNLI-VE} \\ \hline
    & NLVR2 & \multicolumn{2}{|c|}{PathVqa} & \multicolumn{3}{|c||}{SNLI-VE} \\ \hline
    & COCOQA & COCOQA & NLVR2 & COCOQA & NLVR2 & PathVqa \\ \hline
    TAM-CL & \textbf{11.68\%(66.62)} & \textbf{91.28\%(6.58)} & \textbf{95.51\%(50.92)} & \textbf{85.97\%(10.58)} & \textbf{93.92\%(51.36)} &  \textbf{10.19\%(49.69)}\\ 
    ablative $\mathcal{L}_{ikd}$ & 33.42\%(50.07) & 95.04\%(3.73) & 99.58\%(50.07) & 90.15\%(7.41) & 104.24\%(49.26) & 12.98\%(44.29)\\ 
    ablative TAB & 28.87\%(53.68) & 93.65\%(4.79) & 98.69\%(50.24) & 86.68\%(10.05) & 98.39\%(50.30) & 12.94\%(45.96)\\ 
    ablative replay& 31.00\%(51.87) & 98.83\%(0.89) & 95.85\%(50.92) & 99.42\%(0.44) & 95.99\%(50.80) & 64.51\%(19.51)\\
    \hline
\end{tabular}
\caption{The forgetting rate for each task in ablative experiments. For each task sequence, the task names on first row represent the current tasks, the task names on second row represent the previous tasks. Each value means the \textbf{forgetting rate\% (current accuracy)} of previous task after learning the current task.}
\label{tab:ablative}
\end{table*}

\begin{table*}[h]
\centering
\begin{tabular}{||c|c|c|c|c|c|c||}
    \hline
    & $2^{nd}$ task & \multicolumn{2}{|c|}{$3^{rd}$ task} & \multicolumn{3}{|c||}{$4^{th}$ task} \\ \hline
    & $1^{st}$ task & $1^{st}$ task & $2^{nd}$ task & $1^{st}$ task & $2^{nd}$ task  & $3^{rd}$ task \\ \hline
    S-N-C-P & 17.57\%(68.38) & 35.74\%(60.66) & 41.06\%(61.04) & 94.79\%(35.55) & 95.49\%(50.85) &  92.43\%(4.88)\\ 
    P-C-N-S & 5.23\%(50.76) & 6.66\%(49.99) & 53.70\%(19.00) & 11.66\%(47.32) & 53.00\%(19.29) & 0\%(50.86)\\ 
    C-N-P-S & 11.68\%(66.62) & 91.28\%(6.58) & 95.51\%(50.92) & 85.97\%(10.58) & 93.92\%(51.36) & 10.19\%(49.69)\\ 
    N-C-S-P & 24.92\%(64.28) & 25.91\%(64.05) & 29.58\%(43.70) & 103.93\%(49.26) & 92.88\%(4.42) & 95.77\%(35.04)\\
    \hline
\end{tabular}

\caption{The forgetting rate for each task in different task sequences. The leftmost column is the task sequence by the shortcut of different tasks: $\textbf{S}$:SNLI-VE, $\textbf{N}$:NLVR2, $\textbf{C}$:COCOQA, $\textbf{P}$:PathVqa.}
\label{table:analytic}
\end{table*}

To reflect the necessity of each component in our design, ablative experiments are performed on the effect of $\mathcal{L}_{ikd}$ loss function, the size of the memory buffer for experience replay, and using the task attention block, respectively. In ablative experiments, we use the task sequence order COCOQA $\rightarrow$ NLVR2 $\rightarrow$ PathVqa $\rightarrow$ SNLI-VE.

\begin{figure}[t]
\centering
\includegraphics[scale=0.3]{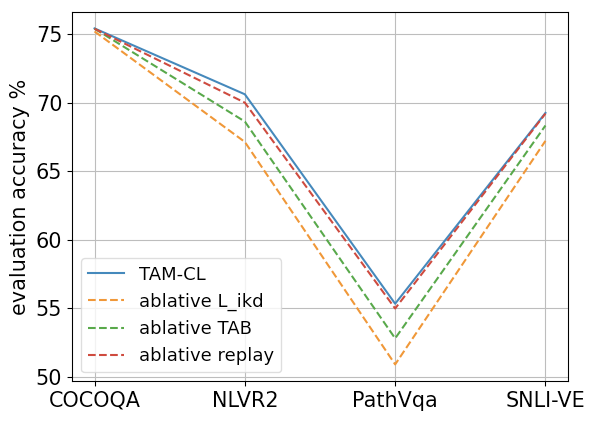}
\caption{Accuracy learning curves for the ablative experiments.}
\label{fig:ablative}
\end{figure}

In the ablative experiment, we choose the full pipeline for TAM-CL as the base-line  and compare the performance of our algorithm in each ablative task.   Table \ref{tab:ablative} presents results for our ablative experiments. We observe that  the full pipeline for TAM-CL leads to the best score at all time-steps in terms of both the forgetting rate and the performance accuracy. These results validate  the necessity of every component in our approach for an optimal performance. We observe that the effect of dropping the $\mathcal{L}_{ikd}$ loss leads to the most significant performance drop which demonstrates that having a teacher model is very helpful for CL. Without $\mathcal{L}_{ikd}$, the model gets the worst forgetting rate on NLVR $\rightarrow$ COCOQA, 33.42\%, PathVqa $\rightarrow$ NLVR2, 99.58\%, and SNLI-VE $\rightarrow$ NLVR2, 104.24\%. Besides   forgetting rate,  the effect of the forward transfer on each task's accuracy also decreases when $\mathcal{L}_{ikd}$ is dropped. In Figure \ref{fig:ablative}, we have presented the accuracy learning curves for the ablative experiments.  We observe that ablating the $\mathcal{L}_{ikd}$ loss terms lead to the worst performance on each of the task. 

Meanwhile, we also observe that when ablating the whole task-attention block, which also includes $\mathcal{L}_{ikd}$, the forgetting rates and accuracy performances are slightly better than only ablating the $\mathcal{L}_{ikd}$. This may look unintuitive but our hypothesis is that the loss term $\mathcal{L}_{ikd}$ is specifically important to train the task-attention layer, as the main continual learning component in the model. In other words, the intermediate-level knowledge distillation in the pre-trained transformer, which is computed by the direct output from the current model and the teacher model as shown in Figure \ref{fig:Big_Diagram}, can help to better constrain the effect of distribution shift on the pretrained parameters and consolidate the knowledge learned for performing  new tasks.

As expected, our training strategy using experience replay is also contributing to the state-of-the-art performance by mitigating catastrophic forgetting of previous tasks. We have provided our ablative results in Table \ref{tab:ablative}. We observe that by ablating experience replay, the forgetting rate of PathVqa $\rightarrow$ COCOQA is 98.83\%, SNLI-VE $\rightarrow$ COCOQA is 99.42\%, and SNLI-VE $\rightarrow$ PathVqa is 64.51\%, which are the highest forgetting rates among the ablative experiments. As COCOQA and PathVqa are the two difficult tasks compared with SNLI-VE and NLVR2, shown in Section \ref{sec:6.1}, the result of ablating experience replay proves its capacity in preventing catastrophic forgetting.
These results demonstrate that the optimal performance of our method stems from using all the three primary ideas.

\subsection{Analytic Experiments}

We provide a second set of analytic experiments to provide better insights about our approach.
To perform analytic experiments, we compare the performance of the TAM-CL architecture on different task sequences and analyze the impact of the task order on catastrophic forgetting. Ideally we would like to develop an algorithm that works well on all task orders. Although in practice we don't control the task order in a CL setting as the tasks are encountered in an order determined by the environment, we study the effect of task order, assuming it is given. We explore the task sequences corresponding to the two extreme cases of task difficulty  orders, mentioned in Section \ref{sec:6.1} which are the two sequences with an increasing and decreasing orders of task difficulties.

\begin{figure}[h]
\centering
\includegraphics[scale=0.3]{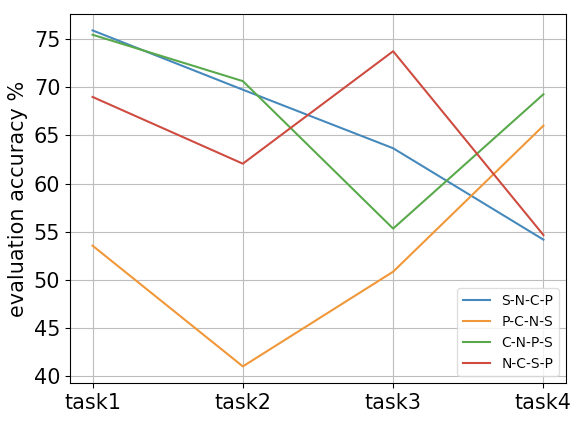}
\caption{Accuracy learning curves for four task sequence orders. }
\label{fig:ablative_accu}
\end{figure}

Table~\ref{table:analytic} presents the forgetting rate results and Figure~\ref{fig:ablative_accu} presents learning curves for all the four orders.
Inspecting results for  the task sequence S-N-C-P, which is the sequence with an increasing difficulty order, we find out that the forgetting rates change in a smoothly increasing pattern, which matches our expectation of task difficulty. In the decreasing difficulty task order, the forgetting rates of each task list is in a relatively low range, compared to the other task sequences, where there are many values in the range of 90\%, and the forgetting rate of NLVR2 after learning SNLI-VE is 0\%. This phenomenon correlate with our hypothesis that the more difficult the task is, the higher the forgetting rates are on the previous tasks. This may be because learning a difficult task can lead to more model updates.

On the other hand, we observe that specifically after learning  PathVqa, the forgetting rates of for previous tasks rise to a high range, around 90\%. However, when the other tasks are learned, the forgetting rates of their previous tasks are mostly below 50\%, except for those learned before PathVqa. We also observe that if PathVqa is the first task, it will affect the accuracy of the following tasks negatively. We note that this trend also exists  for all other CL methods. Our hypothesis is that while datasets such as SNLI-VE, COCOQA, and NLVR2 contain mostly non-field-specific data, PathVqa  is a medical-specified dataset for which all images   are pathological images. When other tasks are sequentially learned, their   distributions are more similar and the effect of domain shift is less significantly because   these dataset are more relevant. However, as PathVqa's dataset is unique in being a medical dataset, updating the model to learn it leads to more forgetting effects on the rest of datasets which are quite different. We conclude that in order to use CL, a priori knowledge about the diversity of observed tasks is helpful.

\section{Conclusions}
We developed an algorithm for multimodal continual learning using transformers based on dynamic model expansion and knowledge distillation. We use a task-attention block which specializes the transformer architecture for a particular tasks using a special task token. Knowledge distillation helps to benefit from knowledge transfer across the tasks for positive forward transfer. We mitigate catastrophic forgetting using experience replay. Our experiments demonstrate that our approach is effective and leads to state-of-the art performance in terms of both forward transfer and catastrophic forgetting.   Our algorithm is a first effort in studying CL in multimodal settings and demonstrates that more explorations in this direction is necessary.

{\small
\bibliographystyle{ieee_fullname}
\bibliography{egbib}
}

\clearpage
\appendix

\section{Implemntation Details}

%
 \subsection{Data Preprocessing}
\begin{table*}[t]
\begin{tabular}{||c|c|c|c|c|c|c||}
    \hline
    \multicolumn{7}{||c||}{NLVR2 $\rightarrow$ COCOQA $\rightarrow$ SNLI-VE $\rightarrow$ PathVqa} \\ \hline
    & COCOQA & \multicolumn{2}{|c|}{SNLI-VE} & \multicolumn{3}{|c||}{PathVqa} \\ \hline
    & NLVR2 & NLVR2 & COCOQA & NLVR2 & COCOQA & SNLI-VE \\ \hline
    TAM-CL & \textbf{24.92\%(64.28)} & \textbf{25.91\%(64.05)} & \textbf{29.58\%(43.70)} & \textbf{96.62\%(50.64)} & \textbf{92.88\%(4.42)} &  \textbf{93.77\%(35.84)}\\ 
    Finetune & 95.52\%(50.87) & 65.83\%(56.66) & 95.96\%(2.67) & 103.97\%(49.20) & 96.98\%(1.99) & 96.83\%(34.63)\\ 
    EWC & 87.49\%(52.45) & 82.88\%(53.25) & 96.35\%(2.37) & 104.39\%(49.14) & 98.55\%(0.94) & 98.15\%(34.09)\\ 
    ER & 46.57\%(60.25) & 36.42\%(62.15) & 33.92\%(42.33) & 104.01\%(49.23) & 95.16\%(3.10) & 94.47\%(35.59)\\
    \hline
    \multicolumn{7}{||c||}{SNLI-VE $\rightarrow$ COCOQA $\rightarrow$ NLVR2 $\rightarrow$ PathVqa} \\ \hline
    & COCOQA & \multicolumn{2}{|c|}{SNLI-VE} & \multicolumn{3}{|c||}{PathVqa} \\ \hline 
    & SNLI-VE & SNLI-VE & COCOQA & SNLI-VE & COCOQA &  NLVR2\\ \hline
    TAM-CL & \textbf{17.11\%(68.60)} &  \textbf{36.64\%(60.29)} & \textbf{66.37\%(22.89)} & \textbf{97.79\%(34.27)} &  \textbf{92.93\%(4.81)} &  95.84\%(\textbf{50.79})\\ 
    Finetune & 85.15\%(39.61) & 99.91\%(33.37) & 96.84\%(2.27) & 98.84\%(33.82) & 96.79\%(2.31) & \textbf{42.16\%}(50.58)\\ 
    EWC & 75.82\%(43.56) & 77.37\%(42.91) & 96.41\%(2.56) & 107.49\%(30.16) & 98.61\%(0.99) & 95.88\%(50.74)\\ 
    ER & 17.89\%(67.79) & 53.46\%(52.86) & 68.82\%(21.41) & 100.99\%(32.92) & 93.58\%(4.62) & 96.07\%(50.74)\\
    \hline
    \multicolumn{7}{||c||}{COCOQA $\rightarrow$ SNLI-VE $\rightarrow$ NLVR2 $\rightarrow$ PathVqa} \\ \hline
    & SNLI-VE & \multicolumn{2}{|c|}{NLVR2} & \multicolumn{3}{|c||}{PathVqa} \\ \hline 
    & COCOQA & COCOQA & SNLI-VE & COCOQA & SNLI-VE &  NLVR2\\ \hline
    TAM-CL & \textbf{6.85\%(69.86)} &  \textbf{20.60\%(59.55)} & \textbf{27.55\%(63.42)} & \textbf{91.86\%(6.11)} &  \textbf{95.65\%(35.14)} &  \textbf{96.40\%(\textbf50.77)}\\ 
    Finetune & 99.21\%(0.60) & 99.57\%(0.32) & 80.10\%(41.68) & 99.03\%(0.73) & 99.51\%(33.54) & 96.95\%(50.56)\\ 
    EWC & 99.38\%(0.46) & 99.30\%(0.53) & 69.81\%(45.93) & 99.13\%(0.65) & 96.33\%(34.86) & 104.83\%(49.13)\\ 
    ER & 14.34\%(64.62) & 57.23\%(32.27) & 45.27\%(56.34) & 95.52\%(3.38) & 99.84\%(33.40) & 103.06\%(49.43)\\
    \hline
\end{tabular}
\caption{Comparative Experiments: the forgetting rate for each task in more task sequences.}
\label{tab:extra_comparative}
\end{table*}
As the tasks, which the model trains on, have completely different dataset, we compress the image from its original size to (384, 640) before sending to the linear projection embedding block. In the self-attention layers of the pre-trained transformer and task-attention block, the image-text feature vector has the shape of (batch, 768). 

Different from tasks such as COCOQA, PathVqa, and SNLI-VE which takes only 1 input image, NLVR2 takes 2 input images with a hypothesis as text input. During the training stage, the text input is combined with one image every time before feeding into the transformer, and concatenate the output from two images as one. Consequently, the output from transformer will be (batch, 1536). However, due to the input size limitation of the task-attention block, we compress the vector, $\mathbf{V}$, from 1536 to 768 by taking the average value of the adjacent element:
\begin{equation}
    \textbf{V}'[i] = \frac{\textbf{V}[i] + \textbf{V}[i+1]}{2}, i=0,2,4,...,1534
\end{equation}
, where $\textbf{V}'$ is the compressed feature vector. We then feed $\textbf{V}'$ into the task-attention block.

\subsection{Hyperparameters for the Experiments}
For all experiments we perform, we use single A100 GPU with batch size of 8. 

For the training epochs, due to the limit of computational force, we only train each task with small epochs: 5 epochs for SNLI-VE, 10 epochs for COCOQA and NLVR2, 15 epochs for PathVQA.

The total training time for each method when performing task sequence COCOQA $\rightarrow$ NLVR2 $\rightarrow$ SNLI-VE $\rightarrow$ PathVqa is 26.9 hours. 


For EWC method, we applied the fisher sample percentage as 0.4, which means 40\% of dataset are collected to build the fisher matrix. During the training stage, we set the EWC weight as 0.1. For Experience Replay method, the sample size of data in memory buffer is 1\%, and the sample frequency is 100, which means for every tasks, we randomly extract 1\% of data and store them into the memory buffer. During the training stage, after 100 batches of current dataset are trained, we randomly select a batch of data from memory buffer and train the model on that batch from a random previous task. For TAM-CL, as the knowledge distillation loss is extremely small compared with other loss, we set the weight for $\mathcal{L}_{ikd}$ as 5000 for all the four tasks. As in TAM-CL, we adopt experience replay as our training strategy, we also set the sample size of memory size as 1\%, and the sample frequency as 100.

\section{Results on Additional Task Orders}

\begin{figure*}
\includegraphics[scale=0.36]{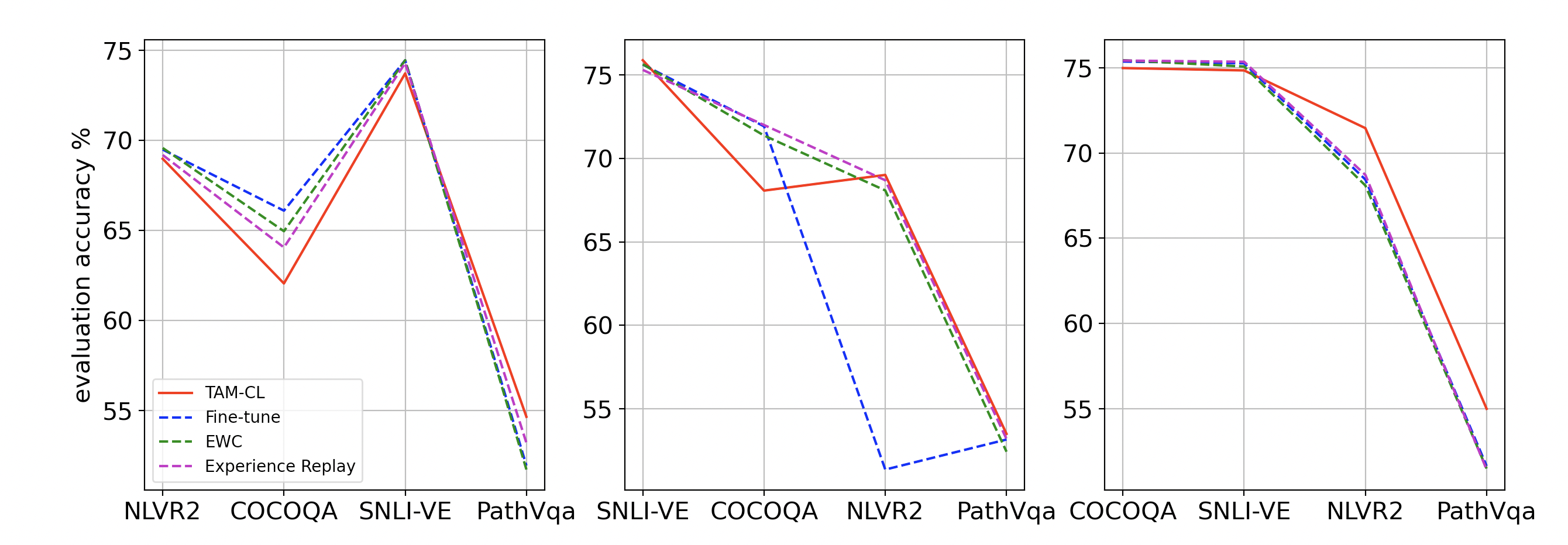}
\caption{The task accuracy of different task sequences}
\label{fig:Extra_Comparative}
\end{figure*}

Due to space limit, additional comparative result on different task sequence are presented here in Table \ref{tab:extra_comparative} and Figure \ref{fig:Extra_Comparative}. Based on the forgetting rates listed on table, we can see TAM-CL still has the leading performance, except the forgetting rate of NLVR2 after learning PathVqa in the second task sequence, where TAM-CL is 95.84\% and Fine-tuning is 42.16\%. However, this doesn't mean TAM-CL is outperformed by Fine-tuning method, as when we focus on the accuracy, after learning task PathVqa, the accuracy for TAM-CL's NLVR2 is 50.79, while the accuracy from Fine-tuning is 50.58, which is slightly lower. This is due to the low performance of NLVR2 on Fine-tuning method with the second task sequence. On the middle figure among Figure \ref{fig:Extra_Comparative}, the accuracy of NLVR2 with Fine-tuning is below 55\%, while all the other methods obtain accuracy around 68\%, which is a huge gap. As the best accuracy of Fine-tuning's NLVR2 is below the others, the forgetting rate of it will be relatively small when the accuracy after learning PathVqa drops to the similar range of other methods.

We also aware that the accuracy of TAM-CL is not always leading among the four methods. However, as all of the differences between the top accuracy and TAM-CL's accuracy are below 5\%, and most of them are even below 1\%, and as our main focus is on the improvement of forgetting rate, we consider the slightly accuracy difference is in an acceptable range. 

\section{Limitations}
Although TAM-CL reaches the state-of-the-art performance, it has its own limitations and we anticipate several future research direction to address these limitation. More specifically:
\begin{itemize}
    \item Currently, as the field-specific dataset, PathVqa, is too diverse from the other datasets, the forgetting rates of previous tasks after learning PathVqa are all above 90\%, which numerically weakens the leading performance of TAM-CL over other methods. In the future work, we will try the experiment with more similar tasks to further prove the leading performance of TAM-CL. Meanwhile, we will also try task sequences with all tasks are from different fields, and analyse the performance of our method.
    
    \item Due to the limit of computational power, our experiments are all performed on single GPU with small batch size. In our future work, we will implement TAM-CL on multi-GPU environment with larger batch size, and analyze its performance compares with running in the single-GPU environment. We have preliminary results that this may  improve our performance further.

    \item In multimodal learning scenario, the modal can not only take multimodal input, but also uni-modal input, by setting the input of other modalities to some constant number. We will further explore TAM-CL's capacity on uni-modal tasks and compare the result with other state-of-the-art methods.
\end{itemize}

\end{document}